\documentclass[acmtog]{acmart}
\AtBeginDocument{%
  
    }

\acmSubmissionID{xxx}

\usepackage{paralist}
\usepackage{placeins}
\definecolor{crimson}{rgb}{0.86, 0.08, 0.24}
\definecolor{gray}{rgb}{0.5,0.5,0.5}
\definecolor{green}{rgb}{0, 0.4, 0}
\definecolor{orange}{rgb}{1, 0.5, 0}
\definecolor{mahogany}{rgb}{0.75, 0.25, 0.0}
\definecolor{purple}{rgb}{0.6, 0, 0.6}
\definecolor{darkgreen}{rgb}{0, 0.4, 0}
\definecolor{frenchblue}{rgb}{0.0, 0.45, 0.73}
\definecolor{blue}{rgb}{0.0, 0.0, 0.65}
\definecolor{red}{rgb}{1,0,0}
\definecolor{yellow}{rgb}{1,1,0}
\definecolor{magenta}{rgb}{1,0,1}
\definecolor{pink}{rgb}{1,0.412,0.706}
\definecolor{cyan}{rgb}{0.25,0.55,0.70}
\definecolor{newgreen}{rgb}{0, 0.6, 0.2}

\newlength\paramargin
\newlength\figmargin
\newlength\figcaptionmargin
\newlength\subfigmargin
\newlength\subsecmargin
\newlength\tabmargin
\newlength\eqmargin

\newlength\presecmargin
\newlength\secmargin

\newlength\rulelength
\def \submission {}
\ifx \submission \undefined
	\newcommand{\yenchi}[1]{{\color{blue}{#1}}}

\else
	\newcommand{\yenchi}[1]{{#1}}

\fi

\begin{document}

\title{AniGS: Bridging Rendering and Diffusion Prior for 3D Scene Animation}

\author{Yen-Chi Cheng}
\authornote{Word was done while intern at Meta.}
\affiliation{%
  \institution{University of Illinois Urbana-Champaign}
  \city{Champaign-Urbana}
  \state{Illinois}
  \country{USA}
}

\author{Chen Gao}
\affiliation{%
  % \institution{Waymo}
  % \city{Bellevue}
  % \state{Washington}
  % \country{USA}
}

\author{Chuhan Chen}
\affiliation{%
  \institution{Carnegie Mellon University}
  \city{Pittsburgh}
  \state{Pennsylvania}
  \country{USA}
}

\author{Tuotuo Li}
\affiliation{%
  \institution{Meta}
  \city{Bellevue}
  \state{Washington}
  \country{USA}
}

\author{Rajvi Shah}
\affiliation{%
  \institution{Meta}
  \city{Bellevue}
  \state{Washington}
  \country{USA}
}

\author{Ayush Saraf}
\affiliation{%
  \institution{Meta}
  \city{Bellevue}
  \state{Washington}
  \country{USA}
}

\author{Changil Kim}
\affiliation{%
  \institution{Meta}
  \city{Bellevue}
  \state{Washington}
  \country{USA}
}

\author{Liangyan Gui}
\affiliation{%
  \institution{University of Illinois Urbana-Champaign}
  \city{Champaign-Urbana}
  \state{Illinois}
  \country{USA}
}

\author{Alexander Schwing}
\affiliation{%
  \institution{University of Illinois Urbana-Champaign}
  \city{Champaign-Urbana}
  \state{Illinois}
  \country{USA}
}

\author{Johannes Kopf}
\affiliation{%
  \institution{Meta}
  \city{Bellevue}
  \state{Washington}
  \country{USA}
}

\author{Hung-Yu Tseng}
% \affiliation{%
%   \institution{}
%   \city{Bellevue}
%   \state{Washington}
%   \country{USA}
% }

\renewcommand{\shortauthors}{Cheng et al.}

\begin{abstract}
Novel view rendering of large and complex reconstructed scenes is becoming increasingly photorealistic.
However, most reconstructions remain static and lack the ambient motion that makes environments immersive.
We present AniGS, a method for scene-level animation of 3D Gaussian Splatting (3DGS) reconstructions that adds subtle, distributed dynamics, e.g., vegetation motion, while preserving rigid structures.
Unlike existing 3D animation techniques which are limited to object-centric subjects or small regions, AniGS is designed for large, cluttered, navigable scenes.
AniGS represents the scene with a canonical 3DGS and models motion using a time-conditioned deformation field. 
To animate the entire scene, we leverage a pretrained video diffusion model and introduce an iterative dataset--model update strategy that progressively expands viewpoint coverage and repeatedly updates camera-fixed training videos using a render-and-refine scheme.
To prevent artifacts from unintended motion in static areas, we further introduce a composed video-to-video refinement scheme that restricts motion to desired regions.
Experiments on five real-world, large-scale outdoor scenes demonstrate that AniGS produces natural ambient dynamics and high-quality novel view videos, enabling more immersive viewing experiences of reconstructed environments.

\end{abstract}



\vspace{-3mm}
\begin{teaserfigure}
  \includegraphics[width=\textwidth]{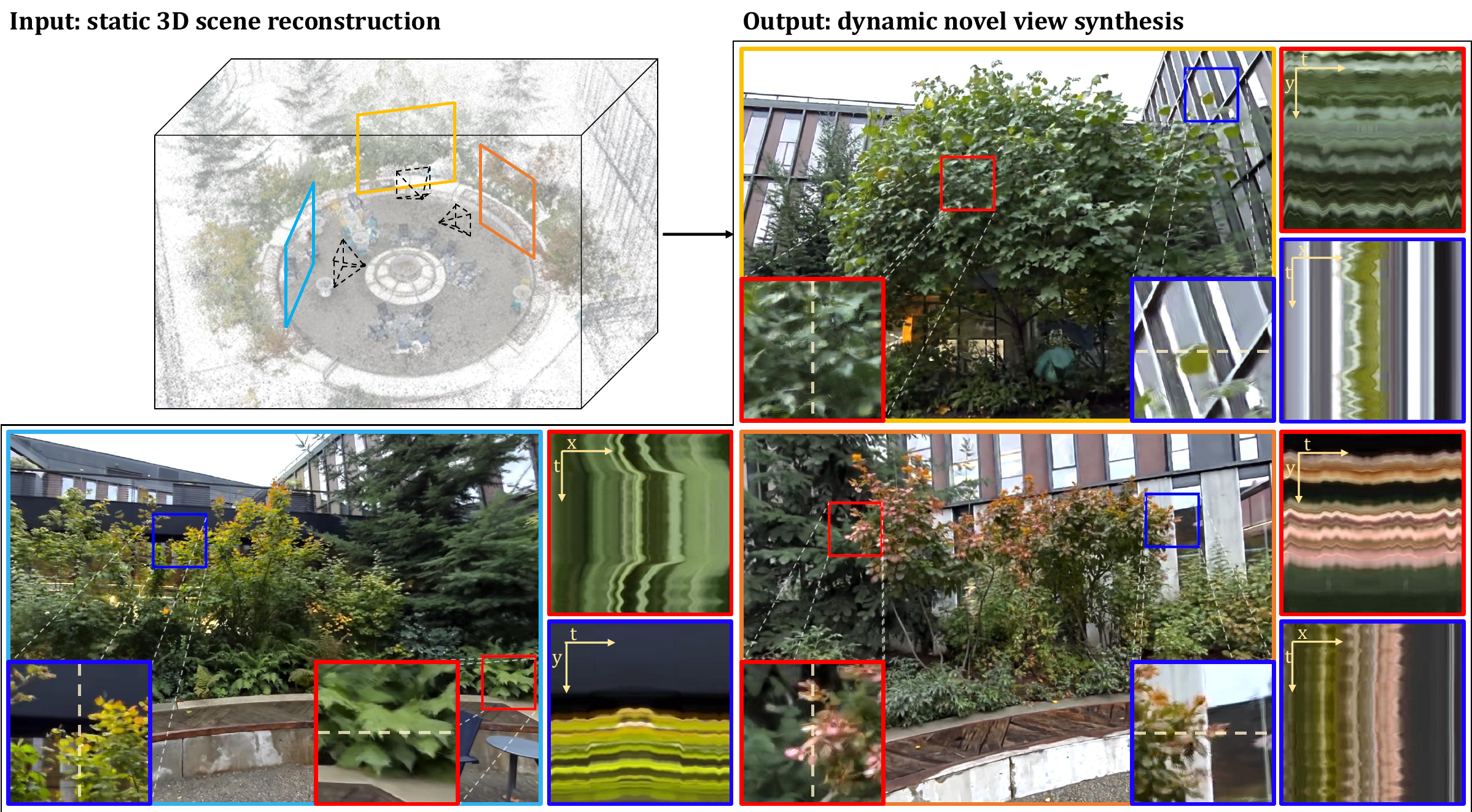}
  \vspace{-7mm}
  \caption{
  \textbf{3D Scene Animation.} Given a static 3D scene (\emph{top-left}) reconstructed with 3D Gaussian Splatting (3DGS), our AniGS method synthesizes an animated 3D scene consisting of ambient motion that supports novel view video rendering. For each example rendered view, we highlight regions of interest and visualize the induced dynamics using space--time slices ($x$/$y$--$t$), showing natural motion in foliage.
  \vspace{-1mm}
  }
  \label{fig:teaser}
\end{teaserfigure}

\received{20 February 2007}
\received[revised]{12 March 2009}
\received[accepted]{5 June 2009}

\maketitle

\vspace{\presecmargin}
\section{Introduction}
\label{sec:intro}

Outdoor environments are inherently dynamic even without moving subjects: wind and subtle material movements cause foliage, grass, and surrounding clutter to move all the time.
This \emph{ambient motion} is critical for an immersive viewing experience.
Although recent advances in 3D Gaussian Splatting (3DGS)~\cite{kerbl20233dgs,yu2024mip_3dgs,lu2024scaffold,huang20242d,held2025triangle,sheng20252dts} enable fast reconstruction of complex 3D scenes and photorealistic novel view rendering, most reconstructed scenes remain \emph{static} and miss the ambient dynamics that make a scene feel alive.
To bridge this gap, rapid progress has been made in animating static 3D assets, especially \emph{object-centric} targets~\cite{zhang2024_physdreamer,jiang2024_animate3d,xie2024physgaussian,wu2025_animateanymesh,kiray2025_promptvfx}.
However, these methods typically focus on a single subject, such as a plant or a toy, and do not scale to large, cluttered, walkable 3D scenes where motion is distributed across many regions.

This work targets \emph{scene-level} animation of 3D Gaussian Splatting (3DGS) reconstructions to provide a more lifelike viewing experience, as illustrated in Figure~\ref{fig:teaser}.
Animating an entire 3D scene is challenging for two reasons.
First, there is no reliable motion prior or supervision that provides temporally consistent motion across multiple viewpoints for the whole scene.
While video diffusion models~\cite{hacohen2024ltx,ali2025cosmos,xing2024dynamicrafter} demonstrate outstanding performance in turning images into high-quality videos, they are not designed to generate multi-view consistent videos. Therefore, they cannot provide temporally coherent motion cues across the full scene.
Second, the animation must preserve static regions, such as walls and floors, which should remain rigid and not drift.
Consider as an example the window view shown in the top-right panel of Fig.~\ref{fig:teaser} (outlined in blue): the window should remain static while the leaves in front sway naturally. The viewing experience degrades noticeably if the static content is not clearly separated from animated regions.

Towards \emph{scene-level} animation, we introduce \textbf{AniGS}, a method that generates an animated 3D scene consisting of ambient motion from a static 3DGS reconstruction.
AniGS uses a canonical 3DGS to represent the scene geometry and appearance, and a time-conditioned deformation field~\cite{shih2024_ambientgs} to model the the synthesized motion.
We leverage a pretrained video diffusion model~\cite{hacohen2024ltx} to provide motion cues.
To overcome the limitation of video diffusion models and animate the whole 3D scene consistently, we design an iterative dataset--model update approach~\cite{haque2023instruct} with an incremental viewpoint expansion strategy.
Specifically, in each iteration, we first add a viewpoint that is near the existing views in the dataset.
We then \emph{update} the camera-fixed videos of all viewpoints in the dataset using a ``render-and-refine'' approach with the video diffusion model.
Finally, the updated videos are used as the training data to optimize the canonical 3DGS and deformation field.
We repeat this dataset--model update cycle until the scene is fully covered and animated.
Furthermore, we observe that the video diffusion model can introduce slight motion in static areas, which leads to obvious artifacts in the final animated scene.
We address this issue with a \emph{composed video-to-video refinement}, which restricts motion to desired regions while preserving the static parts of the scene.

We evaluate the proposed method on five real-world, large-scale, and cluttered outdoor scenes.
Both qualitative and quantitative results show that AniGS 1) adds natural ambient dynamics to appropriate regions and 2) renders high-quality novel view videos.
The results suggest a practical path toward more immersive viewing experiences. We summarize the contributions as follows:
\begin{compactitem}
    \item We propose \textbf{AniGS} for scene-level 3D animation, targeting non-object-centric, cluttered scenes with navigable areas.
    \item We propose an iterative dataset--model update strategy that scales 3D scene animation training to large scenes by progressively generating and incorporating multi-view animated supervision from selected camera viewpoints.
    \item We design a composed video-to-video refinement scheme that steers the video diffusion model to synthesize motion only in desired regions while preserving static content.
\end{compactitem}
\vspace{-3mm}

\vspace{\presecmargin}
\section{Related Works}
\label{sec:related_works}

\begin{figure*}[htbp]
    \centering
    \includegraphics[width=\linewidth]{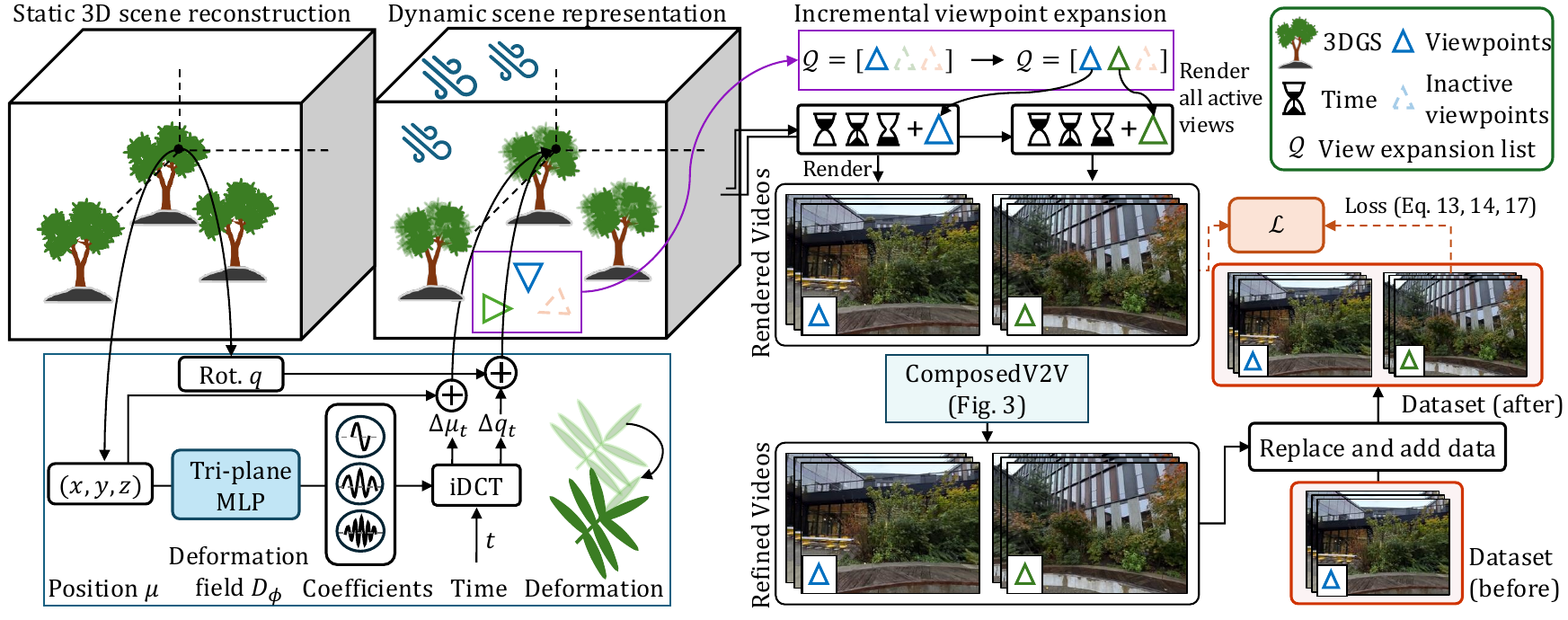}
    \vspace \figmargin
    \caption{
    \textbf{Method overview.} (\emph{left}) AniGS uses a canonical 3DGS to represent scene geometry and appearance, and a time-conditioned deformation field to model the synthesized motion.
    The deformation field predicts discrete cosine transform (DCT) coefficients to offset and rotate each splat in the canonical 3DGS.
    (\emph{right}) To animate the entire 3D scene, we design an iterative dataset--model update scheme with incremental viewpoint expansion strategy. 
    Dataset update: in each iteration, we include an additional viewpoint that has sufficient overlapping with the existing views in the dataset, render camera-fixed videos of all views in the dataset, and refine the videos using the proposed ComposedV2V approach~(Figure~\ref{fig:2_p2_cv2v}).
    Model update: the refined videos are considered as the training data for optimizing the dynamic scene representation.
    We repeat this process iteratively until the scene is fully covered.
    }
    \vspace \figmargin
    \label{fig:2_p1_overview}
\end{figure*}

\noindent\textbf{Dynamic Scene Reconstruction.}
A large body of work studies reconstruction of \emph{time-varying} 3D scenes (4D), often by
introducing temporal degrees of freedom into a 3D representation and optimizing from dynamic observations. Recent Gaussian-splatting-based methods extend 3D
Gaussian Splatting to dynamic settings by learning time-conditioned motion or
deformation for Gaussians, enabling efficient novel view rendering of
dynamic scenes~\cite{wu2024_4dgs,shih2024_ambientgs,jin2025_opt4dgs}. Beyond
purely reconstruction-driven formulations, several works leverage powerful video
generators as priors to recover 4D geometry or enforce multi-view temporal
consistency: Shape-of-Motion lifts motion and structure from a single video to a
4D representation~\cite{wang2024_shapeofmotion}, Geo4D uses video generators to
regularize geometric 4D scene reconstruction~\cite{jiang2025_geo4d}, and
MV-Performer adapts video diffusion models to produce faithful and synchronized
multi-view performer dynamics~\cite{zhi2025_mvperformer}. While effective when
dynamic observations are available, these methods rely on captured
time-varying input (or focus on bounded subjects). In contrast,  our setting starts
from a \emph{canonical static} scene reconstruction and bootstraps scene-wide
animation without scene-level ground-truth dynamic supervision by iteratively
coupling a renderable model with a generative video prior.

\noindent\textbf{4D Generation.}
Recent work has made rapid progress on \emph{generating} dynamic 3D/4D content by combining strong diffusion priors with explicit 3D representations. A number of methods generate 4D objects or scenes from text/images/videos by distilling video diffusion models into Gaussian/NeRF-like representations, producing time-varying geometry and appearance (e.g., DreamGaussian4D~\cite{ren2023_dreamgaussian4d}, 4Dfy~\cite{bahmani2024_4dfy}, and Vivid Dream~\cite{lee2024_vividdream}). More recent approaches further emphasize multi-view and multi-frame consistency for 4D content generation, including autoregressive 4D generation from monocular
videos~\cite{zhu2025_ar4d}, decoupled video diffusion pipelines for single-image
3D/4D scene creation~\cite{sun2025_dimensionx,zhao2025_genxd}, explicit 4D object
generation without score distillation~\cite{sun2025_eg4d}, feedforward
video-to-4D synthesis~\cite{zhang2025_gvfdiffusion}, and multi-view video diffusion models that improve spatial-temporal coherence~\cite{huang2025_mvtokenflow,xie2025_sv4d,yao2025_sv4d2}. Complementary to content generation, DIMO studies diverse 3D motion generation for arbitrary objects~\cite{mou2025_dimo}. Despite impressive results, most of these methods
primarily target \emph{object-centric} content or bounded scenes and rely on
generating content \emph{from scratch} (or from short monocular videos). In contrast, 
our goal is to \emph{animate an entire, cluttered, walkable real-world scene},
starting from a static 3D reconstruction. AniGS leverages a generative
video prior to iteratively bootstrap scene-wide supervision that can be distilled into a renderable 3DGS animation model.%

\noindent\textbf{Animating static 3D Objects and Scenes.}
A related line of work studies generation of motion or animation from an existing \emph{static} 3D asset (mesh, NeRF/3DGS reconstruction) by leveraging generative video priors. Animate3D~\cite{jiang2024_animate3d} and Bringing Objects to Life~\cite{rahamim2024_bringing_objects_to_life} use multi-view/video diffusion guidance to produce view-consistent dynamic renderings and distill them into a time-varying 3D representation. AnimateAnyMesh~\cite{wu2025_animateanymesh} targets feed-forward, text-driven mesh animation, while AKD~\cite{li2025_akd} distills articulated kinematics from diffusion-generated videos. Gaussians2Life~\cite{wimmer2025_gaussians2life} animates a static 3DGS by generating diffusion guidance and lifting the induced 2D motion into 3D Gaussian deformations. PhysDreamer~\cite{xie2024physgaussian,zhang2024_physdreamer} instead distills diffusion-predicted motion into physically grounded material parameters via differentiable simulation. PromptVFX~\cite{kiray2025_promptvfx} explores text-driven fields for animating 3D Gaussians in open-world settings. Despite strong results, most of these methods are \emph{object-centric} or limited to a small region/viewpoint range, and thus do not address globally consistent \emph{scene-level} animation in large, cluttered, walkable environments. Differently, AniGS targets whole-scene animation from a canonical static reconstruction by iteratively bootstrapping supervision with a video diffusion prior.

\vspace{\presecmargin}
\section{Method}
We first describe our problem setting in Sec.~\ref{ssec:problem}. 
We then detail 3D Gaussian Splatting, deformation field, and video diffusion preliminaries in Sec.~\ref{ssec:prelim}. 
Afterwards, we provide an overview of our method in Sec.~\ref{ssec:overview}. 
Subsequently, we discuss how our iterative dataset--model update achieves animation of the entire-scene in Sec.~\ref{ssec:dset_update} and Sec.~\ref{ssec:model_update}. 
Finally, we provide implementation details in Sec.~\ref{ssec:imple_details}.

\vspace{\subsecmargin}
\subsection{Problem definition}
\label{ssec:problem}

Given a \emph{static} scene reconstructed by 3D Gaussian Splatting~(3DGS) $\mathcal{G}^0$, we aim to learn a \emph{scene-level} animated representation that supports novel view rendering of plausible, temporally coherent ambient motion across the entire environment (e.g., trees, grass, flowers, and other foliage), while keeping rigid structures stationary. 
Formally, the output is a time-varying 3DGS $\{\mathcal{G}^t\}_{t=0}^{T}$ that can be rendered from arbitrary viewpoints. 
In AniGS, we do not explicitly predict a set of 3DGS for each time step.
Instead, we represent the animated scene using a canonical static 3DGS and a deformation field, detailed in Section~\ref{ssec:prelim}.

\vspace{\subsecmargin}
\subsection{Preliminaries}
\label{ssec:prelim}

We review the key components used in AniGS: 1) 3D Gaussian Splatting (3DGS) as the canonical static 3D scene representation, 2) a compact dynamic scene parameterization for ambient motion, and 3) a flow-based video diffusion model that we use as a generative prior to introduce dynamics.

\noindent\textbf{3D Gaussian Splatting (3DGS).} 3DGS represents a scene as a set of $N$ 3D Gaussians $\mathcal{G}=\{G_i\}_{i=1}^{N}$. Each Gaussian is defined by a center $\mu\in\mathbb{R}^3$ and a covariance
$\Sigma\in\mathbb{R}^{3\times 3}$:
\begin{equation}
G(x\mid \mu;\Sigma)=\exp\!\left(-\frac{1}{2}(x-\mu)^{\mathsf T}\Sigma^{-1}(x-\mu)\right),
\label{eq:3dgs_gaussian}
\end{equation}
where $\Sigma$ is parameterized by a rotation matrix $R\in\mathbb{R}^{3\times 3}$
and a diagonal scaling matrix $S\in\mathbb{R}^{3\times 3}$ as
\begin{equation}
\Sigma = RSS^{\mathsf T}R^{\mathsf T}.
\label{eq:3dgs_cov}
\end{equation}
In practice, $R$ and $S$ are commonly represented by a unit quaternion $q$ and scaling factors $s\in\mathbb{R}^3$. Besides geometry, each Gaussian carries appearance attributes such as color $c$ and opacity $\alpha$. Rendering is conducted by sorting Gaussians along the ray and accumulating them using the standard over-compositing rule:
\begin{equation}
C = \sum_{i\in\mathcal{N}} c_i \alpha_i \prod_{j=1}^{i-1} (1-\alpha_j),
\label{eq:3dgs_alpha_comp}
\end{equation}
where $\mathcal{N}$ denotes the ordered set of Gaussians contributing to a pixel. The Gaussian parameters (positions, rotations, scales, colors, opacities) are optimized via photometric reconstruction losses between rendered and ground-truth (captured) images. 

In our system, 3DGS serves as the renderable backbone. We initialize AniGS with a \emph{canonical} static reconstruction $\mathcal{G}^0$, representing the 3D scene geometry and appearance at time-zero reference state.

\noindent\textbf{Dynamic Scene Representation for Ambient Motion.}
To model scene-level, ambient dynamics, we parameterize a time-varying deformation on top of a canonical static scene. Inspired by AmbientGS~\cite{shih2024_ambientgs}, we represent per-Gaussian deformations (e.g., translation and rotation updates) with a compact frequency-domain parameterization based on discrete cosine transform (DCT) bases. Specifically, a time-varying scalar deformation $v(t)$ (e.g., $\Delta\mu_t$ along one axis) is
modeled as
\begin{equation}
v(t)=\sqrt{\frac{2}{K+1}}\sum_{k=1}^{K} \phi_{v,k}\cos\!\Big(\frac{\pi}{2T}(2t+1)k\Big),
\label{eq:dct_motion}
\end{equation}
where $\phi_{v,k}$ are learnable coefficients and $K$ controls the basis size. This representation is storage-efficient and generalize smoothly over time; at any timestamp $t$ we recover $\Delta \mu_i(t)$ and $\Delta r_i(t)$ (via iDCT) and apply them to the canonical Gaussians before rendering.

\noindent\textbf{Flow-based Video Diffusion Model.} 
We use a flow-based (rectified-flow) video diffusion formulation in latent space to provide the motion cues. Given a clean latent $z_0$ (e.g., encoded from a video), we interpolate between $z_0$ and Gaussian noise $\epsilon\sim\mathcal{N}(0,I)$ via
\begin{equation}
z_\tau = (1-\tau)z_0 + \tau\epsilon,\quad \tau\in[0,1].
\label{eq:rf_forward}
\end{equation}
Instead of predicting $\epsilon$, rectified-flow training typically predicts the \emph{velocity} $v=\epsilon-z_0$ to balance learning difficulty across diffusion timesteps $\tau$. During inference, starting from pure noise $z_1$, the model integrates the reverse-time ODE using Euler updates:
\begin{equation}
z_{\tau-\Delta \tau} = z_\tau - \Delta \tau\, v_{\theta}(z_\tau,\tau).
\label{eq:rf_euler}
\end{equation}
In practice, the objective can also be described as mapping a noisy latent $z_{\tau_i}$ to the expected clean latent $z_0$ (i.e., $v_\theta(z_{\tau_i},\tau_i)\approx z_0$), with $z_{\tau_i}$ constructed as in Eq.~\eqref{eq:rf_forward}. 
We use pretrained LTX~\cite{hacohen2024ltx} in AniGS.

\begin{figure}[t]
    \centering
    \includegraphics[width=\linewidth]{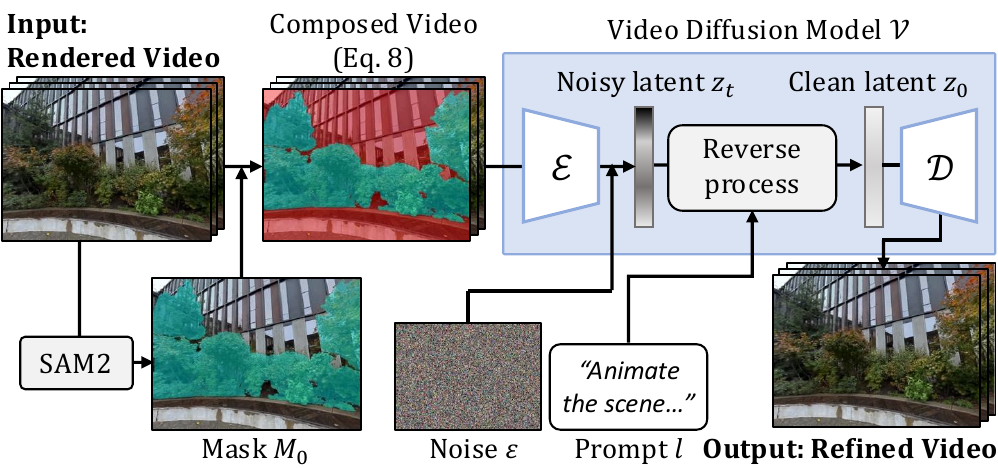}
    \vspace \figmargin
    \caption{\textbf{Composed Video-to-Video (\textsc{ComposedV2V}).}
    Slight motion introduced by the video diffusion model in stationary regions can be amplified over the iterative dataset--model update process.
    We therefore propose ComposedV2V to \emph{re-stabilize} the static region in each dataset update.
    Given a rendered (camera-fixed) video, we estimate the dynamic mask from the first frame, and create the composed video by \emph{copying} the static pixels of the first frame to all subsequent frames~(Eq.~\eqref{eq:compose_render}). Intuitively, this enforces stability in stationary areas. Conditioning on the text prompt $l$ and noise $\epsilon$, we apply the video diffusion model $\mathcal{V}$ to obtain the refined video.
    }
    \label{fig:2_p2_cv2v}
\end{figure}

\vspace{\subsecmargin}
\subsection{Overview of AniGS}
\label{ssec:overview}

We now provide an overview of the proposed method (Fig.~\ref{fig:2_p1_overview}). Our goal is to animate an entire outdoor scene starting from a canonical static reconstruction $\mathcal{G}^0$ (Sec.~\ref{ssec:problem}). A key obstacle is the lack of suitable ground-truth supervision: it is impractical to obtain videos that are temporally consistent \emph{for the whole scene}. A single camera has a limited field of view and cannot observe all regions at once, while capturing scene-wide, temporally synchronized motion would require multiple calibrated cameras recording the environment concurrently, which is infeasible for typical casual capture.

To address this obstacle, AniGS follows an iterative \emph{dataset--model update} routine that combines a renderable representation with a generative prior from a pretrained (flow-based) video diffusion model (Sec.~\ref{ssec:prelim}). Specifically, we parameterize scene motion with a time-conditioned deformation field $D_{\phi}$ over the canonical Gaussians, producing $\mathcal{G}^{t}=D_{\phi}(\mathcal{G}^{0}, t)$ and rendering frames from arbitrary viewpoints via $\hat{I}_{p,t}=\mathrm{Render}(\mathcal{G}^{t},p)$. Instead of learning $D_{\phi}$ from ground-truth dynamic labels, we periodically update the training dataset using the current model renderings and diffusion-based video-to-video refinement, and then optimize $D_{\phi}$ using these updated videos.

Starting from $\mathcal{G}^0$, we repeatedly alternate between 1) dataset update (Sec.~\ref{ssec:dset_update}) and 2) model update (Sec.~\ref{ssec:model_update}). This loop bootstraps temporally coherent supervision for the \emph{entire scene} and yields a final animated scene model.

\noindent\textbf{Dataset Update.} In each iteration $k$, we first perform an \emph{incremental viewpoint expansion} by adding a viewpoint from a set of candidate camera views to the training dataset. Note that this incremental process gradually expand scene coverage. Then for each view $p^{(j)}$ in the training dataset, we use the current renderable representation $\mathcal{G}^{t}=D_{\phi}(\mathcal{G}^0,t)$ to render a camera-fixed video $R^{(j)}_{0:T}$:
\begin{equation}
    R^{(j)}_{0:T}=\{\mathrm{Render}(\mathcal{G}^{t}, p^{(j)})\}_{t=0}^{T}.
\end{equation}
Finally, we refine the video using a pretrained video diffusion model via \emph{composed video-to-video refinement} (Section~\ref{ssec:dset_update}).
The refined videos at all views in the training dataset $\{(p^{(j)}, A^{(j)}_{0:T})\}$ form the new training dataset for the next iteration.

\noindent\textbf{Model Update.} With the updated dataset
$\{(p^{(j)}, A^{(j)}_{0:T})\}$, we optimize the renderable model for several iterations before the next dataset update. Concretely, we update the time-conditioned deformation field $D_{\phi}$ (and the center parameters $\mu$ in the canonical 3DGS $\mathcal{G}^0$), encouraging the rendered videos to match the videos in the newly updated dataset. To stabilize training and prevent drift, we additionally regularize the dynamics in the canonical space, encouraging the animated Gaussians to remain consistent with the canonical reconstruction when appropriate. We provide the details of our formulation in Sec.~\ref{ssec:model_update}.

By repeating iterative dataset--model update for multiple rounds, AniGS progressively expands supervision coverage and learns a globally consistent animated 3DGS.

\vspace{\subsecmargin}
\subsection{Dataset Update}
\label{ssec:dset_update}
Learning scene-wide animation in a single shot is impractical in our setting. First, optimizing deformations for all Gaussians and all viewpoints at once is computationally prohibitive and often leads to out-of-memory errors. Second, dense ground-truth dynamic supervision with full scene coverage is unavailable under casual capture, as it would require many synchronized cameras. To address both concerns, we progressively bootstrap supervision by \emph{iteratively updating the training dataset} (See Fig.~\ref{fig:2_p1_overview}).

\noindent\textbf{Incremental viewpoint expansion.}
We grow the dataset by adding one view at a time. We first select $N$ camera locations $\{p_{(i)}\}_{i=1}^{N}$ based on segmentation map of regions of interest (e.g., trees, flowers) and \yenchi{use farthest-point sampling to expand the views for the full scene coverage.}
These cameras are ordered counter-clockwise. For each location $p_{(i)}$, we consider three viewing directions $\psi\in\{0^\circ,90^\circ,-90^\circ\}$ and render each direction with a fixed field-of-view (FOV=$95^\circ$) to ensure overlap among neighboring directions.

We maintain a queue $\mathcal{Q}$ (\texttt{views\_update\_list}) of views to be added. Starting from the first camera and the forward direction, we append $(p_{(1)},0^\circ)$ into $\mathcal{Q}$. We then iterate until convergence: at each dataset update, we pop the next view $(p,\psi)$ from $\mathcal{Q}$, render a clip from the current model at this view, and use our \textsc{ComposedV2V} module (Sec.~\ref{ssec:dset_update}) to generate a refined animated target for that view. After a view is processed, we enqueue the next direction at the same camera following a fixed order $0^\circ \rightarrow 90^\circ \rightarrow -90^\circ$; once all three directions of $p_{(i)}$ are processed, we proceed to the next camera $p_{(i+1)}$. In practice, we update the dataset periodically so that newly added views are generated using the latest state of the model.

\noindent\textbf{Composed-Video-to-Video Refinement.}
A naive dataset update strategy directly runs a video diffusion model on each rendered view and uses the generated clips as supervision. In practice, this often fails for two reasons. First, independently generated clips across different viewing directions are typically temporally inconsistent (e.g., producing different motion phases for the same content in overlapping regions), which provides conflicting supervision and prevents the dynamic Gaussians from converging. Second, the dataset is periodically updated throughout training; if the generated targets are not aligned with the current state of the renderable model, the supervision can drift and destabilize optimization. To mitigate these issues, we propose a \emph{Composed Video-to-Video Refinement} (Fig.~\ref{fig:2_p2_cv2v}) that tightly couples the diffusion-based generation with the current model rendering while explicitly encouraging stable static regions.

Given a selected view $p^{(j)}$ and timesteps $\mathbf{T}=(0,1,\ldots,T)$, we first render the current model to obtain a clip $R^{(j)}_{0:T}=\{R^{(j)}_{t}\}_{t=0}^{T}$, where $R^{(j)}_{t}=\mathrm{Render}(\mathcal{G}^{t},p^{(j)})$. We then construct a \emph{render mask} ${M}_{0}^{(j)}\in\{0,1\}^{H\times W}$ from the first frame $R^{(j)}_{0}$ using SAM2~\cite{ravi2024sam2}. Using this mask, we form a \emph{composed} conditioning clip $\bar{R}^{(j)}_{0:T}$ by pasting the canonical static region from $t{=}0$ to all frames:
\begin{equation}
    \bar{R}^{(j)}_{t} = \left(\mathbf{1}-M_{0}^{(j)}\right)\odot R^{(j)}_{0} + M_{0}^{(j)}\odot R^{(j)}_{t},\qquad \forall t\in\mathbf{T}. \label{eq:compose_render}
\end{equation}
Intuitively, $\bar{R}^{(j)}_{0:T}$ provides a stable camera shot and enforces static floor/background, while leaving the region of interest available for motion synthesis. 

Finally, we feed $\bar{R}^{(j)}_{0:T}$ to a pretrained flow-based video diffusion model with prompt ${p}^{(j)}$ and noise $\boldsymbol{\epsilon}^{(j)}$ to obtain the refined animated target clip
\begin{equation}
A^{(j)}_{0:T}
=\textsc{ComposedV2V}\!\left(
R^{(j)}_{0:T},\, {M_0}^{(j)},\, {l},\, \boldsymbol{\epsilon}^{(j)}
\right).
\label{eq:composed_v2v_refine}
\end{equation}
Here, \textsc{ComposedV2V} encapsulates the above composition together with the diffusion-based video-to-video refinement. Specifically, we use a rectified-flow (flow-matching) formulation in the latent space. Let $z^{(j)}_{0:T}=\mathcal{E}(\bar{R}^{(j)}_{0:T})$ denote the composed clip encoded by a video VAE encoder $\mathcal{E}(\cdot)$. We construct a noisy latent at diffusion timestep $\tau\in[0,1]$ by linearly interpolating between the clean latent and Gaussian noise:
\begin{equation}
z^{(j)}_{\tau} = (1-\tau)\,z^{(j)}_{0:T} + \tau\,\boldsymbol{\epsilon}^{(j)},\qquad
\boldsymbol{\epsilon}^{(j)}\sim\mathcal{N}(0,\mathbf{I}),
\label{eq:rf_forward_v2v}
\end{equation}
and condition the flow network on the prompt $l$ via cross-attention. The model predicts the rectified-flow velocity $v_{\theta}(z^{(j)}_{\tau},\tau,l)$, which defines a reverse-time ODE. Starting from pure noise $z^{(j)}_{1}=\boldsymbol{\epsilon}^{(j)}$, we integrate the reverse process with Euler steps:
\begin{equation}
z^{(j)}_{\tau-\Delta \tau} = z^{(j)}_{\tau} - \Delta \tau\; v_{\theta}\!\left(z^{(j)}_{\tau},\tau,{l}\right),
\qquad \tau:1\rightarrow 0,
\label{eq:rf_reverse_euler_v2v}
\end{equation}
to obtain the refined clean latent $\hat{z}^{(j)}_{0:T}$ at $\tau=0$. Finally, we decode it back to pixel space with the VAE decoder $\mathcal{D}(\cdot)$:
\begin{equation}
A^{(j)}_{0:T}=\mathcal{D}\!\left(\hat{z}^{(j)}_{0:T}\right).
\label{eq:rf_decode_v2v}
\end{equation}
In summary, \textsc{ComposedV2V} takes the composed conditioning clip $\bar{R}^{(j)}_{0:T}$ (derived from $R^{(j)}_{0:T}$ and ${M_0}^{(j)}$), adds noise as detailed in Eq.~\eqref{eq:rf_forward_v2v}, and runs the reverse rectified-flow process conditioned on prompt $l$ to produce a temporally coherent, refined video $A^{(j)}_{0:T}$.

\vspace{\subsecmargin}
\subsection{Model Update}
\label{ssec:model_update}

Given the updated dataset produced by the method described in Sec.~\ref{ssec:dset_update}, we optimize the renderable representation using objectives which compare the rendered clips and the pseudo-target animations. For a training view $p_i$ and video timestep $t$, we rasterize the deformed Gaussians $\mathcal{G}^{t}=D_{\phi}(\mathcal{G}^0,t)$ to render the frame $R_{t}^{p_i}=\mathrm{Render}(\mathcal{G}^{t},p_i)$, and supervise it with the refined target $A_{t}^{p_i}$ using a standard photometric loss:
\begin{equation}
\mathcal{L}_{\mathrm{ani}}
=
\left\|R_{t}^{p_i}-A_{t}^{p_i}\right\|_{1}
+\bigl(1-\mathrm{SSIM}(R_{t}^{p_i},A_{t}^{p_i})\bigr).
\end{equation}
To stabilize optimization and reduce drift artifacts (e.g., floating splats and blur), we additionally regularize the model in the canonical space. Concretely, we enforce the canonical rendering to remain close to the static appearance at $t=0$ by comparing the rendered canonical frame $R_{0}^{p_i}$ with the static reference image $I_{0}^{p_i}$:
\begin{equation}
\mathcal{L}_{\mathrm{cano}}
=
\left\|R_{0}^{p_i}-I_{0}^{p_i}\right\|_{1}
+\bigl(1-\mathrm{SSIM}(R_{0}^{p_i},I_{0}^{p_i})\bigr).
\end{equation}
\yenchi{To further improve animation quality, we additionally apply a score distillation sampling (SDS) loss. For a training camera $p_i$, we sample a starting timestep $t$ and render a short clip of $4m{+}1$ frames,
\(
R^{p_i}_{t:t+4m}
=
\{R^{p_i}_{t},R^{p_i}_{t+1},\ldots,R^{p_i}_{t+4m}\},
\)
where $m$ is an integer. We encode this clip into the latent space of the video diffusion model using the VAE encoder $\mathcal{E}(\cdot)$:
\begin{equation}
z^{p_i}_{0}=\mathcal{E}\!\left(R^{p_i}_{t:t+4m}\right).
\end{equation}
We then sample a diffusion timestep $\tau\sim\mathcal{U}(0,1)$ and Gaussian noise $\epsilon\sim\mathcal{N}(0,\mathbf{I})$, and construct the noisy latent following the same rectified-flow formulation used in Sec.~\ref{ssec:dset_update}:
\begin{equation}
z^{p_i}_{\tau}=(1-\tau)\,z^{p_i}_{0}+\tau\,\epsilon.
\end{equation}
Given the text prompt $l$, the pretrained DiT predicts the rectified-flow velocity $v_{\theta}(z^{p_i}_{\tau},\tau,l)$. Since the target velocity under this formulation is $\epsilon-z^{p_i}_{0}$, we define the SDS loss as
\begin{equation}
\mathcal{L}_{\mathrm{SDS}}
=
\mathbb{E}_{p_i,t,\tau,\epsilon}
\left[
w(\tau)\,
\left\|
v_{\theta}(z^{p_i}_{\tau},\tau,l)
-\left(\epsilon-z^{p_i}_{0}\right)
\right\|_2^2
\right],
\end{equation}
where $w(\tau)$ is a timestep-dependent weighting function. This loss encourages the rendered animation to follow the motion prior of the pretrained video diffusion model while remaining consistent with our renderable scene representation. In practice, due to memory constraints, we back-propagate gradients only through one sampled frame among the $4m{+}1$ frames in the rendered clip.}
Hence, our training objective is $\mathcal{L}=\mathcal{L}_{\mathrm{ani}}+\lambda_{\mathrm{cano}}\mathcal{L}_{\mathrm{cano}} + \lambda_{\mathrm{SDS}} \mathcal{L}_{\mathrm{SDS}}$.

In terms of the learnable parameters, we find that optimizing only the deformation field $D_{\phi}$ while freezing all canonical Gaussian attributes yields suboptimal results. Instead, we additionally fine-tune the \emph{canonical positions} $\{\mu_i^0\}$ to better align geometry with the diffusion-refined targets, while keeping other attributes (rotation, scale, opacity, and color) fixed. This parameterization improves training stability and facilitates scene-wide animation.

\vspace{\subsecmargin}
\subsection{Implementation Details}
\label{ssec:imple_details}

\noindent\textbf{Static reconstruction.}
We initialize AniGS with a canonical static 3DGS $\mathcal{G}^0$. We optimize $\mathcal{G}^0$ using AdamW with a learning rate of $1\times10^{-4}$. We train for 30k iterations with around 1.6 hours to converge.

\noindent\textbf{Animation training.}
We start from $\mathcal{G}^0$ and learn scene dynamics by optimizing the time-conditioned deformation field $D_{\phi}$ together with the \emph{canonical Gaussian positions} $\{\mu_i^0\}$, while keeping other canonical attributes (e.g., rotation, opacity, and appearance) fixed. We use $N{=}4$ camera locations for the iterative dataset update. For the generative prior, we use LTX-Video~\cite{hacohen2024ltx} as the pretrained video diffusion model. We run the diffusion model with 8 sampling steps to generate $T{=}121$ frames at a resolution of $1024\times1408$. During training, we update the dataset via \textsc{ComposedV2V} every $2500$ optimization steps per viewing direction, and train AniGS for 70k iterations in total. The canonical mask for \textsc{ComposedV2V} is obtained with SAM2. Training time in this stage is about 7.3 hours with 1 GPU.

\vspace{\presecmargin}
\section{Experiment}

In this section, we evaluate AniGS on a dataset of five challenging outdoor scenes captured in casual settings, each containing large-scale, cluttered vegetation and a wide walkable regions. We compare against representative 3D scene animation baselines, Gaussians2Life~\cite{wimmer2025_gaussians2life} and PhysDreamer~\cite{zhang2024_physdreamer}. We first describe the dataset and baselines in Sec.~\ref{ssec:exp_dataset}-~\ref{ssec:exp_baselines}. We then report both qualitative results and comparisons (Fig.~\ref{fig:3.1.qual_ours}-\ref{fig:3.2.qual_comp}), and quantitatively measure animation realism and quality with metrics and a user study in Sec.~\ref{ssec:exp_quan}-\ref{ssec:exp_ablation}.

\vspace{\subsecmargin}
\subsection{Dataset}
\label{ssec:exp_dataset}
We collect a dataset consisting of five large-scale, cluttered outdoor scenes: \textit{Fireplace}, \textit{Garden}, \textit{Flowers}, \textit{Forest}, and \textit{Trail}. These captures cover diverse vegetation and clutter (e.g., trees, flowers, plants, and bushes) over a walkable region, and are thus well-suited for evaluating scene-wide animation. Each scene contains approximately 1.5k--2.0k monocular images, and we estimate the camera poses and the sparse point cloud using COLMAP~\cite{schonberger2016colmap}. \yenchi{We additionally evaluate on the public benchmark DL3DV~\cite{ling2024dl3dv}. Specifically, we select outdoor scenes from the dataset using the coarse category annotation ``Nature \& Outdoors.'' The selected scenes include \textit{Gazebo}, \textit{Courtyard}, \textit{Bush}, \textit{Playground}, and \textit{Park}.}

\vspace{\subsecmargin}
\subsection{Baselines}
\label{ssec:exp_baselines}
We compare our AniGS with two representative diffusion-based 3D animation methods: \textbf{Gaussians2Life}~\cite{wimmer2025_gaussians2life} and \textbf{PhysDreamer}~\cite{zhang2024_physdreamer}. Gaussians2Life animates a static 3DGS by generating multi-view diffusion guidance and lifting the induced 2D motion into 3D via tracked points and depth alignment to drive Gaussian deformations. PhysDreamer instead distills diffusion-predicted motion into physically grounded dynamics by optimizing material parameters of a differentiable simulator (MPM) so that simulated renderings match generated video clips. For a fair comparison, we provide both baselines with the same pretrained static reconstruction for each scene, and use their official implementations with a shared training view due to their object-centric setup.

\vspace{\subsecmargin}

\subsection{Evaluation Metrics}

\noindent\textbf{Metrics.} We evaluate animation quality using Fr\'echet Video Distance (FVD)~\cite{unterthiner2018fvd} and Fr\'echet Inception Distance (FID)~\cite{heusel2017fid}. We compute FVD with an I3D backbone~\cite{carreira2017i3d}; all videos are resized and center-cropped to $256{\times}256$ before feature extraction. Since ground-truth scene-level dynamic videos are unavailable in our setting, we measure FVD against a \emph{reference} video distribution generated by pretrained video diffusion models. For a fair comparison and to reduce dependence on a specific generator, we use three diffusion models--LTX~\cite{hacohen2024ltx}, Cosmos~\cite{ali2025cosmos}, and DynamiCrafter~\cite{xing2024dynamicrafter}--to generate reference videos, reporting the resulting scores as \textit{FVD-LTX}, \textit{FVD-Cosmos}, and \textit{FVD-DynamiCrafter}, respectively. For each method, we generate 250 clips when computing FVD. We compute FID by treating diffusion-generated frames pooled across all scenes as the reference distribution, and evaluating FID on the frames rendered by each method for each scene.

\noindent\textbf{User study.} We further conduct a user study to assess visual quality and motion realism. We follow a two-alternative forced choice (2AFC) protocol: for each question, we render two clips from the \emph{same} view, showing AniGS and a randomly selected baseline side-by-side. The left-right ordering is randomly permuted to mitigate side bias. Participants answer two questions per pair: (1) which clip has better visual quality, and (2) which clip has more realistic motion. We report the fraction of votes preferring each method.

\vspace{\subsecmargin}

\subsection{Qualitative evaluation}
\label{ssec:exp_qual}

\begin{figure*}[htbp]
    \centering
    \includegraphics[width=0.85\linewidth]{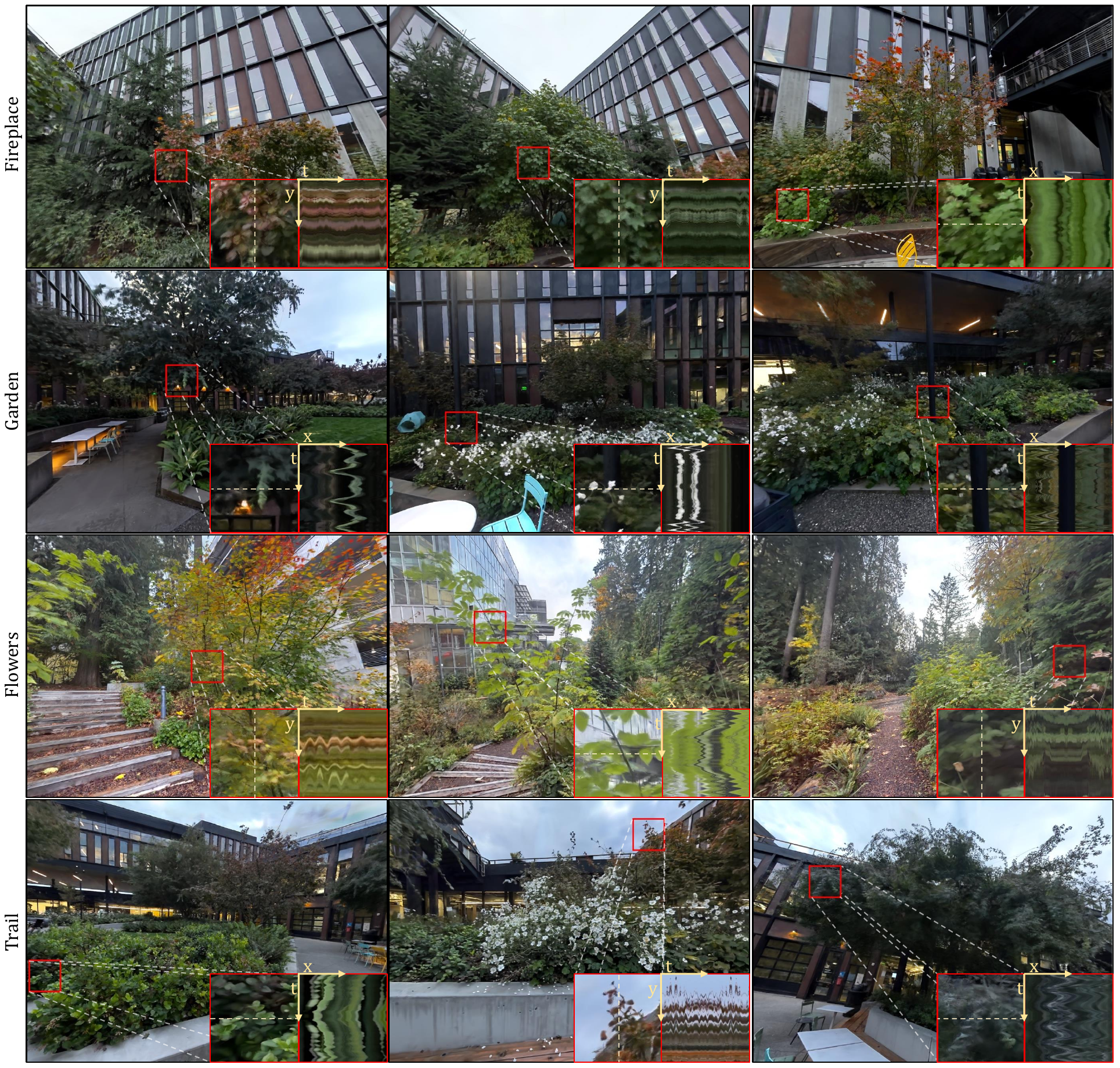}
    \vspace \figmargin
    \caption{
    \textbf{Qualitative results}. We present the novel view rendering results on various scenes, and highlight the synthesized dynamics using space-time slices.
    }
    \vspace{-3mm}
    \label{fig:3.1.qual_ours}
\end{figure*}

\begin{figure*}[htbp]
    \centering
    \includegraphics[width=0.6\linewidth]{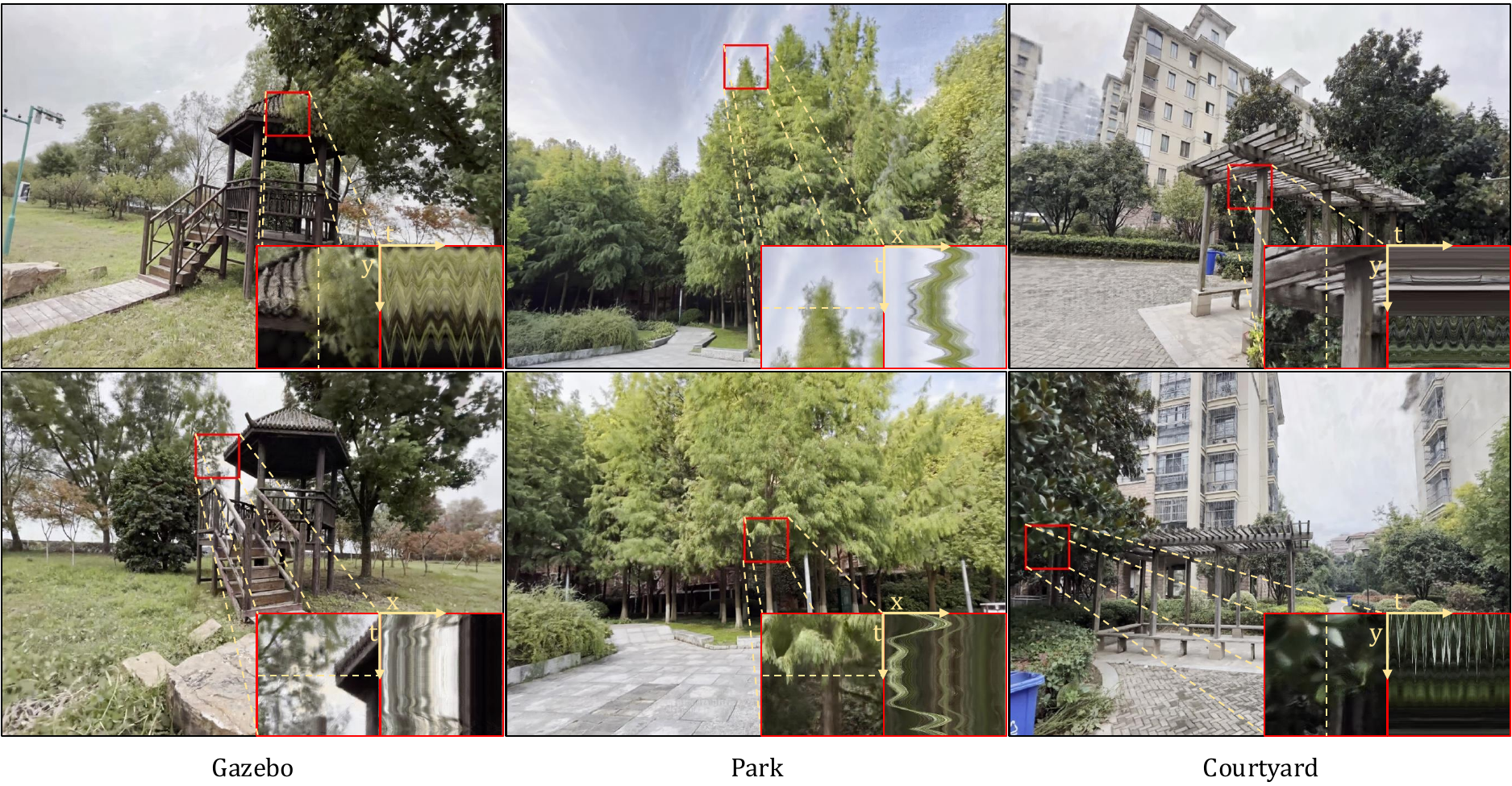}
    \vspace \figmargin
    \caption{
    \textbf{Qualitative results on DL3DV}. \yenchi{We present the novel view rendering results on scenes from DL3DV and highlight the animation.}
    }
    \vspace \figmargin
    \label{fig:3.5.qual_ours_dl3dv}
\end{figure*}

We visualize qualitative results of AniGS in Fig.~\ref{fig:3.1.qual_ours} and Fig.~\ref{fig:3.5.qual_ours_dl3dv}. For each scene, we render animations from multiple camera locations to demonstrate scene-level coverage, and for each rendering we show 1) the first frame and 2) an $x$-$t$/$y$-$t$ slice of a highlighted patch to illustrate temporal evolution. The results show that AniGS produces natural, temporally coherent motion on cluttered dynamic regions (e.g., leaves and flowers), while preserving appearance for structures such as buildings and other background elements.

\begin{figure*}[htbp]
    \centering
    \includegraphics[width=0.8\linewidth]{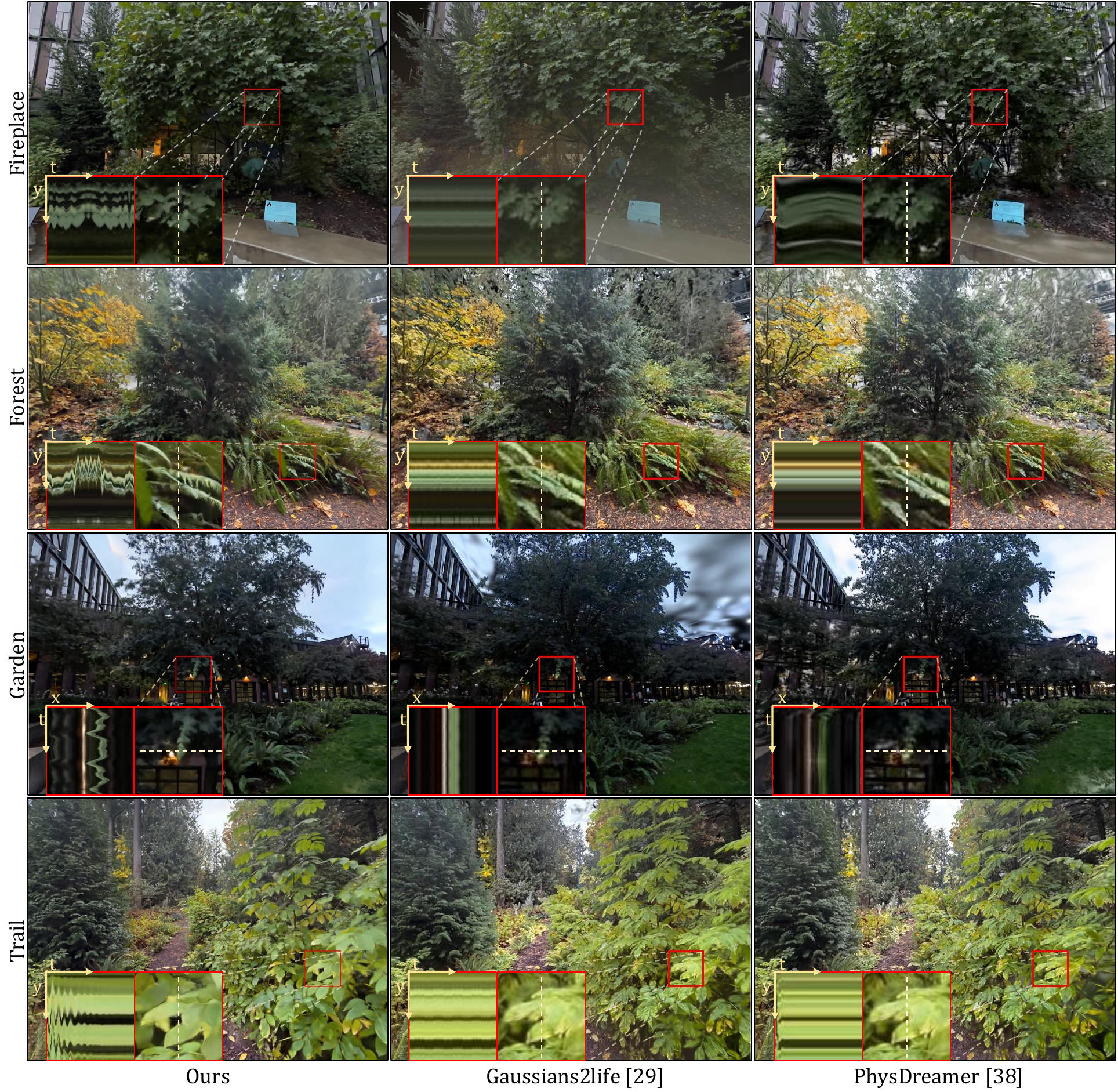}
    \vspace \figmargin
    \caption{
    \textbf{Qualitative comparisons}. We present the novel view rendering results produced by different approaches: ours, Gaussians2life~\cite{wimmer2025_gaussians2life} and PhysDreamer~\cite{zhang2024_physdreamer}.
    The synthesized dynamics is shown using space-time slices.
    }
    \vspace \figmargin
    \label{fig:3.2.qual_comp}
\end{figure*}

We further compare AniGS with Gaussians2Life and PhysDreamer in Fig.~\ref{fig:3.2.qual_comp}. AniGS consistently yields clearer renderings and more realistic motion patterns (e.g., oscillatory leaf/flower motion resembling wind-driven dynamics) without introducing noticeable background drift. In contrast, Gaussians2Life often produces locally or spatially fragmented motion that is confined to small regions. PhysDreamer often produces less realistic dynamics because it induces near-global shifting of the entire scene rather than generating foliage-specific motion. These highlight the advantage of our method to produce motion on regions of interest while maintaining a stable canonical structure for the remainder of the scene. Please check the supplementary for the rendered videos.

\begin{figure}[t]
    \centering
    \includegraphics[width=0.7\linewidth]{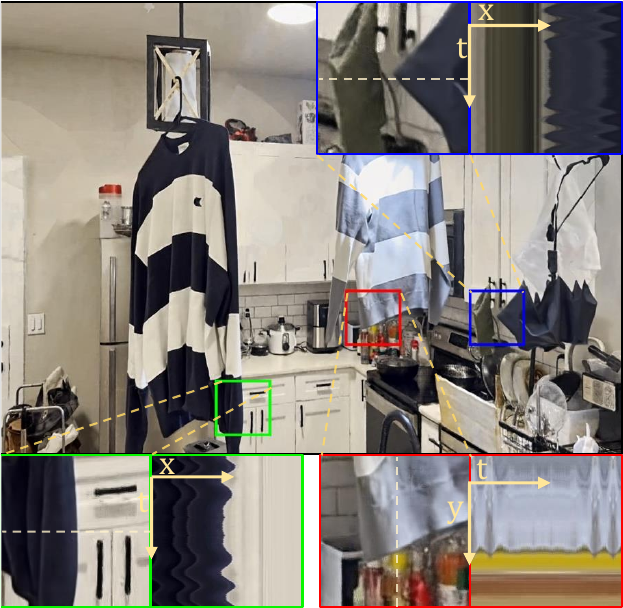}
    \caption{
    \textbf{Diverse animation result.} \yenchi{AniGS can generalize beyond vegetation to diverse about such as clothes, umbrella, and plastic bags.}
    }
    \vspace \figmargin
    \label{fig:3.6.diverse}
\end{figure}

\yenchi{Finally, we show a diverse scene animation result in Fig.~\ref{fig:3.6.diverse}, where AniGS animates an indoor scene containing clothes, an umbrella, and plastic bags. This result demonstrates that AniGS can generalize beyond outdoor vegetation and work in different scenarios with diverse animatable objects.}

\vspace{\subsecmargin}
\subsection{Quantitative evaluation}
\label{ssec:exp_quan}
We report quantitative results \yenchi{on our collected dataset using FVD~\cite{unterthiner2018fvd} as shown in Tab.~\ref{tab:quan.comp.fvd}. We also report FVD on DL3DV in Tab.~\ref{tab:quan.dl3dv.fvd}}. Across all five scenes, AniGS achieves consistently lower FVD than both baselines under three different diffusion reference distributions (LTX/Cosmos/DynamiCrafter), indicating more realistic and temporally coherent motion.
\yenchi{AniGS also performs favorably compared to the baselines on DL3DV scenes as shown in Tab.~\ref{tab:quan.dl3dv.fvd}}

\yenchi{We also evaluate the multiview consistency of the animated videos in Tab.~\ref{tab:quan.3d.consist}. We conduct a 3D reconstruction experiment to verify that the generated multiview motion is 3D consistent. Specifically, we select a novel timestep ($t{=}21$), render multiview images from $50\%$ of the training cameras, train a 3DGS reconstruction using these rendered images, and evaluate PSNR on the remaining validation cameras at Tab.~\ref{tab:quan.3d.consist}. We include a 3DGS trained on the original captures as a reference. The results show that the learned animation is multiview-consistent.}

Finally, we conduct a user study in Tab.~\ref{tab:quan.user}. Under the Two-Alternative Forced Choice (2AFC), participants prefer AniGS over Gaussians2Life and PhysDreamer for both motion realism and visual quality in all scenes.

\begin{table}[t]
\caption{
    \textbf{Quantitative evaluation (FVD)}.
    We report the FVD ($\downarrow$) score computed using various video diffusion models: LTX, Cosmos, and DynamiCrafter. The best performance is in \textbf{bold}.
}
\vspace{\tabmargin}
\centering
\setlength{\tabcolsep}{0.25em}
\footnotesize
\begin{tabular}{l c c c c c c} 
    \toprule
    \textbf{FVD-LTX} $\downarrow$ & Fireplace & Garden & Flowers & Forest & Trail & Avg. \\
    \midrule
    Ours & \textbf{374.49} & \textbf{403.89} & \textbf{537.14} & \textbf{407.86} & \textbf{356.08} & \textbf{415.89} \\
    Gaussians2life & 1127.78 & 876.60 & 1154.95 & 605.64 & 745.19 & 902.03 \\
    PhysDreamer & 943.56 & 957.12 & 1078.05 & 550.37 & 668.20 & 839.46 \\
    \midrule
    \textbf{FVD-Cosmos} $\downarrow$ \\
    \midrule
    Ours & \textbf{224.03} & \textbf{254.32} & \textbf{307.70} & \textbf{183.23} & \textbf{165.68} & \textbf{226.99} \\
    Gaussians2life & 739.64 & 653.49 & 899.55 & 381.19 & 476.39 & 630.05 \\
    PhysDreamer & 677.38 & 531.39 & 946.43 & 496.17 & 511.22 & 632.52 \\
    \midrule
    \textbf{FVD-DynamiCrafter} $\downarrow$ \\
    \midrule
    Ours & \textbf{476.17} & \textbf{465.13 }& \textbf{485.96} & \textbf{550.58} & \textbf{499.54} & \textbf{495.48} \\
    Gaussians2life & 668.16 & 772.07 & 1044.03 & 782.01 & 854.38 & 824.13 \\
    PhysDreamer & 691.74 & 855.37 & 947.86 & 1096.62 & 709.83 & 860.28 \\
    \bottomrule
\end{tabular}
\vspace{\tabmargin}
\label{tab:quan.comp.fvd}
\end{table}

\begin{table}[t]
\caption{
    \textbf{Quantitative evaluation on DL3DV.}
    \yenchi{We report FVD-LTX ($\downarrow$) on outdoor scenes from DL3DV~\cite{ling2024dl3dv}. The best performance is shown in \textbf{bold}.}
}
\vspace{\tabmargin}
\centering
\setlength{\tabcolsep}{0.25em}
\footnotesize
\begin{tabular}{l c c c c c c} 
    \toprule
    \textbf{FVD-LTX} $\downarrow$ & Gazebo & Park & Bush & Playground & Courtyard & Avg. \\
    \midrule
    Ours & \textbf{413.59} & \textbf{387.04} & \textbf{310.19} & \textbf{452.92} & \textbf{343.66} & \textbf{381.48} \\
    Gaussians2life & 945.42 & 839.34  & 735.03 & 815.52 & 630.99 & 793.26 \\
    PhysDreamer & 881.19 & 762.17 & 651.94 & 790.86  & 986.31 & 814.49 \\
    \bottomrule
\end{tabular}
\vspace{\tabmargin}
\label{tab:quan.dl3dv.fvd}
\end{table}

\begin{table}[t]
\caption{
    \yenchi{\textbf{Multiview consistency of generated videos.}
    We evaluate multiview consistency by training a 3DGS from generated multiview images at a novel timestep and reporting PSNR.
    }
}
\vspace{\tabmargin}
\centering
\setlength{\tabcolsep}{0.25em}
\footnotesize
\begin{tabular}{l c c c c} 
    \toprule
    \textbf{PSNR} $\uparrow$ & Fireplace & Garden & Gazebo & Courtyard  \\
    \midrule
    Original capture & 27.40 & 28.77 & 26.29 & 28.35 \\
    Our Novel Time & 25.94 & 27.13 & 24.41 & 26.87 \\
    \bottomrule
\end{tabular}
\vspace{\tabmargin}
\label{tab:quan.3d.consist}
\end{table}

\vspace{\subsecmargin}
\subsection{Ablation study}
\label{ssec:exp_ablation}

We validate key components of AniGS in Tab.~\ref{tab:ablation}. Disabling the \emph{dataset--model update} (i.e., generating pseudo-training videos only once) degrades quality significantly because the supervision quickly becomes misaligned with the evolving renderer, making optimization harder to converge. Removing the \emph{incremental view expansion} (i.e., having all viewpoints in the training dataset at the beginning) further destabilizes training, as the model loses the curriculum effect that gradually expands scene coverage. Turning off \textsc{ComposedV2V} also hurts results, since independently generated clips are less temporally consistent across views and introduce conflicting supervision. Finally, we find that parameter updates must be carefully controlled: optimizing all canonical Gaussian attributes is prone to blur and drift, while optimizing only the deformation field is restricted by static geometric bias; allowing canonical \emph{positions} to be fine-tuned provides a better canonical alignment so the deformation focuses on residual time-varying motion.

\vspace{\subsecmargin}

\begin{table}[t]
\caption{
    \textbf{User study.} We report user preference rates for AniGS compared with other methods, assessed in terms of motion realism and visual quality.
}
\vspace{\tabmargin}
\centering
\setlength{\tabcolsep}{0.25em}
\small
\begin{tabular}{l c c c c c} 
    \toprule
    \textbf{Motion realism} $\uparrow$ & Fireplace & Garden & Flowers & Forest & Trail \\
    \midrule
    Ours vs GS2life & 91.7 \% & 83.3 \% & 81.5 \% & 78.3 \% & 83.3 \% \\
    Ours vs PhysDreamer & 89.6 \% & 90.5 \% & 85.7 \% & 90.8 \% & 91.7 \% \\
    \midrule
    \textbf{Visual quality} $\uparrow$ \\
    \midrule
    Ours vs GS2life & 90.2 \% & 82.6 \% & 80.7 \% & 74.5 \% & 86.3 \% \\
    Ours vs PhysDreamer & 96.0 \% & 79.1 \% & 83.5 \% & 79.2 \% & 87.9 \% \\
    \bottomrule
\end{tabular}
\vspace{\tabmargin}
\label{tab:quan.user}
\end{table}

\begin{table}[H]
\caption{
    \textbf{Ablation study}. We report FVD ($\downarrow$) scores computed using the LTX video diffusion model. The best performance is in \textbf{bold}.
}
\vspace{\tabmargin}
\centering
\setlength{\tabcolsep}{0.25em}
\footnotesize
\begin{tabular}{l c c c c c c } 
    \toprule
    FVD-LTX $\downarrow$ & Fireplace & Garden & Flowers & Forest & Trail  & Avg. \\
    \midrule
    w/o dataset--model update & 533.95 & 529.67 & 653.92 & 518.72 & 431.49 & 533.55 \\ %
    w/o viewpoint expansion & 511.46 & 498.50 & 597.16 & 489.12 & 410.87 & 501.42 \\
    w/o \textsc{ComposedV2V} & 420.77 & 425.20 & 556.29 & 433.17 & 373.02 & 441.69 \\ %
    w/ opt. all param. & 473.14 & 445.93 & 607.83 & 458.36 & 396.55 & 476.36 \\
    w/ opt. deformation only & 450.22 & 439.16 & 610.53 & 449.83 & 402.14 & 470.38 \\
    Ours (full model) & \textbf{374.49} & \textbf{403.89} & \textbf{537.14} & \textbf{407.86}& \textbf{356.08} & \textbf{415.89} \\
    \bottomrule
\end{tabular}
\vspace{\tabmargin}
\label{tab:ablation}
\end{table}

\vspace{\presecmargin}
\section{Conclusion and Limitation}

\noindent\textbf{Conclusion.} We introduced AniGS, which introduces scene-wide dynamics to a static 3DGS reconstruction by coupling a renderable model with a pretrained video diffusion prior. AniGS learns a time-conditioned deformation field without ground-truth dynamic supervision via an iterative dataset--model update loop that expands view coverage, refines renderings into temporally coherent pseudo training data, and optimizes the model with canonical regularization for free-viewpoint animation.

\begin{figure}[t]
    \centering
    \includegraphics[width=0.8\linewidth]{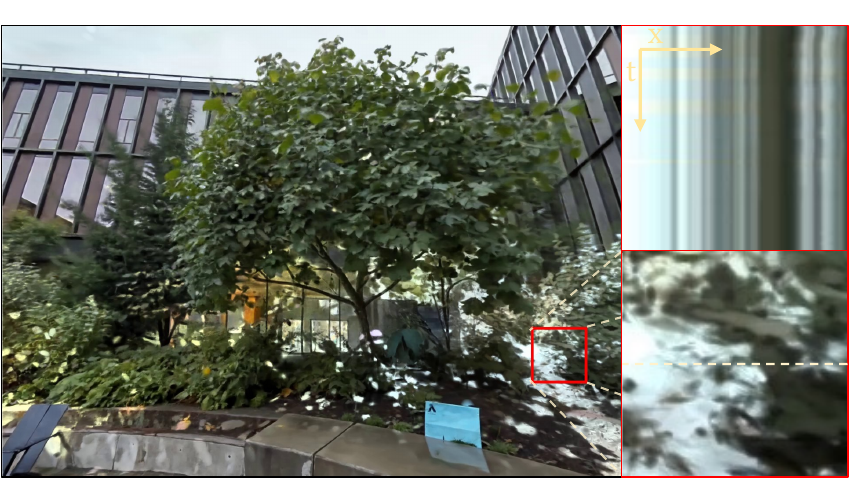}
    \caption{
    \textbf{Limitation example.}
    AniGS cannot synthesize large, non-reversible dynamics.
    We show an example of the strong time-lapse effects of lighting changes.
    AniGS failed to model the global changes produced by the video diffusion model in this case and produce noticeable artifacts.
    }
    \vspace \figmargin
    \label{fig:3.4.failure}
\end{figure}

\noindent\textbf{Limitation.} AniGS does not synthesize \emph{state changes} or other large, non-reversible dynamics. In particular, if the diffusion-refined targets contain events such as leaves falling, branches breaking, or strong time-lapse effects (e.g., day-to-night or seasonal transitions) that induce drastic global appearance/tone changes, the current deformation-based representation and training objective struggle to explain such large motion or global shifts. An example of strong time-lapse effects is presented in Figure~\ref{fig:3.4.failure}.
We leave modeling these to future work.%

\bibliographystyle{ACM-Reference-Format}
\bibliography{citation}

\FloatBarrier

\FloatBarrier

\end{document}